\documentclass{article}

\PassOptionsToPackage{numbers,compress}{natbib}
    \usepackage[preprint]{neurips_2023}

\usepackage[utf8]{inputenc} % allow utf-8 input
\usepackage[T1]{fontenc}    % use 8-bit T1 fonts
\usepackage{hyperref}       % hyperlinks
\usepackage{url}            % simple URL typesetting
\usepackage{booktabs}       % professional-quality tables
\usepackage{amsmath,amsfonts}       % blackboard math symbols
\usepackage{nicefrac}       % compact symbols for 1/2, etc.
\usepackage{microtype}      % microtypography
\usepackage{xcolor}         % colors
\usepackage{algorithmic}
\usepackage{algorithm}
\usepackage{array}
\usepackage{textcomp}
\usepackage{stfloats}
\usepackage{url}
\usepackage{verbatim}
\usepackage{graphicx}
\usepackage{booktabs}
\usepackage{relsize}
\usepackage{placeins}
\usepackage{caption}
\usepackage{subcaption}
\usepackage{multirow}
\newtheorem{theorem}{Theorem}
\newtheorem{lemma}{Lemma}

\title{Targeted collapse regularized autoencoder for anomaly detection: black hole at the center}

\author{%
  Amin Ghafourian\\
  Department of Mechanical\\and Aerospace Engineering\\
  University of California, Davis\\
  Davis, CA 95616 \\
  \texttt{aghafourian@ucdavis.edu} \\
  \And
  Huanyi Shui \\
  Ford Motor Company \\
  Dearborn, MI 48126 \\
  \texttt{hshui@ford.com} \\
  \And
  Devesh Upadhyay \\
  Ford Motor Company \\
  Dearborn, MI 48126 \\
  \texttt{dupadhya@ford.com} \\
  \AND
  Rajesh Gupta \\
  Ford Motor Company \\
  Dearborn, MI 48126 \\
  \texttt{rgupta39@ford.com} \\
  \And
  Dimitar Filev \\
  Ford Motor Company \\
  Dearborn, MI 48126 \\
  \texttt{dimitar.filev@gmail.com} \\
  \And
  Iman Soltani\thanks{Corresponding author} \\
  Department of Mechanical\\and Aerospace Engineering\\
  University of California, Davis\\
  Davis, CA 95616 \\
  \texttt{isoltani@ucdavis.edu} \\
}

\begin{document}

\maketitle

\begin{abstract}
Autoencoders have been extensively used in the development of recent anomaly detection techniques. The premise of their application is based on the notion that after training the autoencoder on normal training data, anomalous inputs will exhibit a significant reconstruction error. Consequently, this enables a clear differentiation between normal and anomalous samples. In practice, however, it is observed that autoencoders can generalize beyond the normal class and achieve a small reconstruction error on some of the anomalous samples. To improve the performance, various techniques propose additional components and more sophisticated training procedures. In this work, we propose a remarkably straightforward alternative: instead of adding neural network components, involved computations, and cumbersome training, we complement the reconstruction loss with a computationally light term that regulates the norm of representations in the latent space. The simplicity of our approach minimizes the requirement for hyperparameter tuning and customization for new applications which, paired with its permissive data modality constraint, enhances the potential for successful adoption across a broad range of applications. We test the method on various visual and tabular benchmarks and demonstrate that the technique matches and frequently outperforms more complex alternatives. We further demonstrate that implementing this idea in the context of state-of-the-art methods can further improve their performance. We also provide a theoretical analysis and numerical simulations that help demonstrate the underlying process that unfolds during training and how it helps with anomaly detection. This mitigates the black-box nature of autoencoder-based anomaly detection algorithms and offers an avenue for further investigation of advantages, fail cases, and potential new directions.
\end{abstract}
\setcounter{footnote}{0}

\section{Introduction}
{A}{nomaly} detection is the task of identifying instances of data that significantly deviate from normal\footnote{The term ``normal'' in this text indicates ``not anomalous.''} observations. The task has many important applications including fault detection~\cite{purarjomandlangrudi2014data,yan2023hybrid,liang2020sparse,zhang2022anomaly,fan2020defective,chen2023adaptive}, medical diagnosis~\cite{fernando2021deep,han2021madgan,kascenas2022denoising,tschuchnig2022anomaly}, fraud detection~\cite{pourhabibi2020fraud,pang2019deep,pang2019deviation,zheng2019one,hilal2022financial}, performance and throughput anomaly detection and bottleneck identification\cite{ibidunmoye2015performance,cherkasova2009automated,chen2023hierarchical}, and network intrusion detection~\cite{ahmed2016survey,al2018deep,dutta2020deep,aygun2017network,wang2020cloud,lokman2019deep,naseer2018enhanced,yang2020real}, among others. Considering that anomalous samples are rare and can often highly vary, anomaly detection is typically done through learning regularities present in normal data. Subsequently, samples that reside far outside the normal class representation are regarded as anomalous.

Deep autoencoders~\cite{hinton2006reducing,bengio2006greedy} are powerful tools to learn unsupervised representations of high dimensional data and have been central to many algorithms among various modern techniques for anomaly detection~\cite{sakurada2014anomaly,zhou2017anomaly,gong2019memorizing,gao2022tsmae}. The premise for using autoencoders in this context is that when trained on normal samples, they fail to effectively reconstruct anomalous samples as by definition they lie outside the training distribution. It is therefore possible to devise an anomaly scoring based on the reconstruction error. This premise, however, does not always hold and there indeed can be cases where an anomalous input is well reconstructed~\cite{perera2019ocgan,zong2018deep}.

Various strategies have been proposed in the literature to improve the performance of autoencoders by including additional components and more sophisticated training procedures. Some examples of this include adding a memory module~\cite{gong2019memorizing,gao2022tsmae,yang2021memory}, visual and latent discriminators~\cite{yang2021memory,perera2019ocgan}, classifier~\cite{perera2019ocgan}, adversarial training~\cite{yang2021memory,perera2019ocgan}, and negative mining~\cite{perera2019ocgan}. The added modules and involved training procedures can lead to increased costs associated with the utilization of such models, potentially making them less accessible to various applications and more challenging to troubleshoot.

In this work, we adopt two principles and aim to shed light on the mechanisms by which they enhance the utility of autoencoders in anomaly detection. First, we promote feature compactness in the latent space, ensuring that normal samples exhibit similar representations; second, we require feature descriptiveness so that the latent representation retains relevant information for effective separation between normal and anomalous samples. Based on these principles, we propose a regularization that penalizes the norm of latent representations. This is a simple and computationally efficient modification, especially given the lower dimensionality of the latent space. We will show that this simple adjustment makes the anomaly detection performance of an encoder-decoder architecture match or surpass that of many deep techniques that incorporate the complex modifications mentioned earlier. We also theoretically analyze a simplified case and use the results to offer an explanation regarding the effects of norm minimization on learning dynamics and how it might help with anomaly detection. We further demonstrate that by virtue of promoting compact representations, the proposed regularization can enhance the performance of other techniques that rely on the separation of learned normal/anomalous latent representations.

The rest of the paper is organized as follows: in Section~\ref{Related Work} we discuss related work on anomaly detection and autoencoder regularization. Section~\ref{Methods} discusses the proposed methodology and theoretical analysis, as well as empirical demonstrations for a simplified case. In Section~\ref{Results} we present anomaly detection results on benchmark datasets as well as ablation studies. Section~\ref{Conclusion} concludes the paper.

\section{Related Work}\label{Related Work}

Regularized autoencoders have shown improvements over their unregularized counterparts, resulting in better performance on various applications due to the advantages they introduce~\cite{alain2014regularized,guo2016deep,makhzani2013k,vincent2010stacked,rifai2011contractive,kingma2013auto}. Sparse autoencoders~\cite{makhzani2013k} regularize the network to encourage sparsity in hidden layers, which serves as an information bottleneck. Denoising autoencoders~\cite{vincent2010stacked} promote robustness to corrupted data and better expressive features by training the model to reconstruct a denoised version of the noisy input. Expanding on a similar intuition, contractive autoencoders~\cite{rifai2011contractive} promote feature robustness to variations in the neighborhood of training instances by penalizing the norm of the Jacobian matrix of the encoder's activations with respect to input samples. Variational autoencoders~\cite{kingma2013auto} regularize the training by introducing the prior that additionaly incentivizes samples to follow a standard Gaussian distribution in latent space and thus learns an approximation of data distribution parameters rather than arbitrary functions for encoding and reconstruction. These variants of autoencoders have been extensively used in applications of anomaly detection~\cite{sakurada2014anomaly,zhou2017anomaly} including fault detection and industrial health monitoring~\cite{yan2023hybrid,liang2020sparse,zhang2022anomaly,fan2020defective}, network intrusion, anomaly, and cyber-attack detection~\cite{al2018deep,dutta2020deep,aygun2017network,wang2020cloud,lokman2019deep,naseer2018enhanced,yang2020real,nguyen2019gee}, video anomaly detection~\cite{narasimhan2018dynamic}, medical diagnosis~\cite{kascenas2022denoising}, geochemical exploration~\cite{xiong2021robust}, and high-energy physics searches~\cite{finke2021autoencoders}. 

To improve the performance of autoencoders in anomaly detection, some recent works propose using external memory to learn and record prototypical normal patterns~\cite{gong2019memorizing,gao2022tsmae,yang2021memory}. The encoding is then used as a query to read the most relevant items from the memory based on an attention mechanism. The combined memory items are then used for reconstruction.~\cite{yang2021memory} demonstrates that by incorporating bidirectional generative adversarial networks (BiGANs)~\cite{donahue2016adversarial} with adversarial training and adding a cycle-consistency loss, memory units improve and tend to lie on the boundary of the convex hull of normal data encodings.

One-Class GAN (OCGAN)~\cite{perera2019ocgan} focuses on ensuring that anomalous samples are poorly reconstructed by bounding the latent space and ensuring various regions of it will only represent examples of the normal class and only normal class samples can be generated from it through the decoder. To do this, the algorithm bounds the latent space and incorporates adversarial training with latent and visual discriminators, as well as a classifier that is used to ensure generated samples belong to the normal class. In addition,~\cite{perera2019ocgan} leverages negative mining to actively seek and improve upon regions of latent space that produce samples of poor quality.

Besides these works that use adversarial training mainly to aid with the reconstruction-based anomaly score and autoencoder performance, there are many works based on GANs and adversarial training that explicitly use the discriminator output to complement or entirely substitute the reconstruction error~\cite{schlegl2017unsupervised,zhao2016energy,zenati2018adversarially,akcay2019ganomaly}. There are also techniques that attempt to directly learn the anomaly score. This can, for instance, be based on training an ordinal regression neural network that gives different scores to sample pairs given the pair composition~\cite{pang2019deep}. Differently, likelihood-based techniques operate on the premise that normal and anomalous instances respectively correspond to high and low-probability events and therefore the model is optimized to, for instance, obtain high probability for training samples using a softmax or a noise contrastive estimate (NCE) objective~\cite{gutmann2010noise,chen2016entity}.

\section{Methods}\label{Methods}

\subsection{Problem Statement}

Semi-supervised anomaly detection (sometimes also referred to as unsupervised anomaly detection) can be formally defined as the following: given a dataset of $m-$dimensional normal samples $\mathcal{X}=\{x^{(i)}\}_{i=1}^n$ (i.e.~samples that are believed to belong to the distribution representing the normal class), obtain an anomaly scoring function $a(\cdot):\mathbb{R}^m\mapsto\mathbb{R}$ that assigns larger scores to anomalous samples and smaller scores to normal samples.

Deep methods incorporate neural networks to parameterize various functions within this process. Reconstruction-based methods mainly use autoencoders to parameterize and learn intermediate variable calculations (i.e. compression and reconstruction functions) toward evaluating the scoring function. A typical reconstruction-based anomaly score formulation for query sample $x_q$ is
\begin{equation}
    a(x_q) = \lVert\hat{x}_q-x_q\rVert_2;\quad\hat{x}_q=De(En(x_q;\theta_e);\theta_d),
\end{equation}
where $En(\cdot;\theta_e)$ and $De(\cdot;\theta_d)$ are the encoder and decoder, respectively parameterized by $\theta_e$ and $\theta_d$. It is then expected that normal samples (i.e.~samples from the same distribution as training) incur smaller reconstruction error compared to anomalous, out-of-distribution samples. Prior research~\cite{perera2019ocgan,zong2018deep}, however, shows that the assumptions on reconstruction quality might be dubious and the effectiveness of anomaly scores can vary depending on the specific type of anomaly. If the anomalous sample shares some critical features with normal samples, it may still be well reconstructed.

\subsection{Targeted Collapse Regularized Autoencoders}

\begin{figure}[!t]
\centering
    \includegraphics[width=0.8\linewidth]{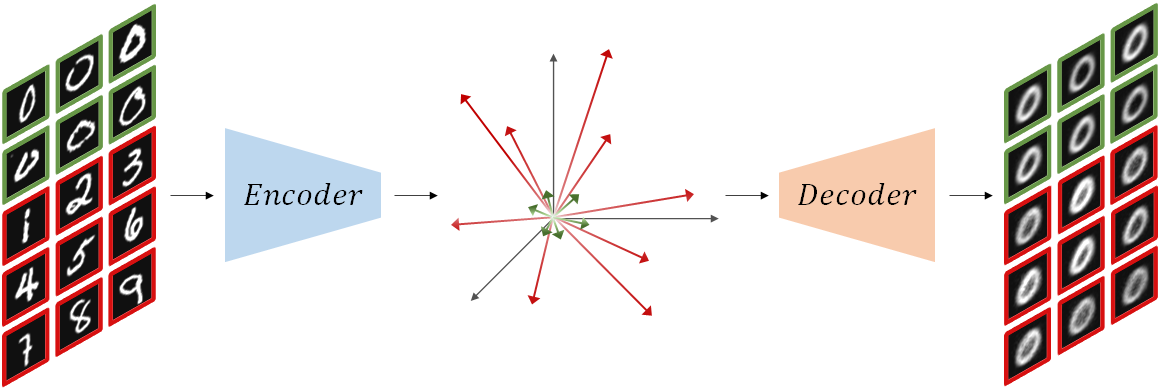}
\caption{Overview of anomaly detection with Toll. Normal and anomalous samples and their associated latent representations and reconstructions are respectively represented by green and red outlines and vectors. During training, in addition to minimizing reconstruction error, latent representation norms are also minimized, with hyperparameter $\beta$ specifying the trade-off between the two terms. To use the trained model for anomaly detection, the overall loss associated with a sample is used as the anomaly score. It is expected that anomalous samples incur larger values of combined reconstruction error and latent representation norm as specified by $\beta$. (Figure best viewed in color.)}
\label{fig:Toll}
\end{figure}

Considering the limitation of autoencoders, we propose an approach which we call Toll (\underline{T}argeted c\underline{oll}apse) that incorporates a simple adjustment applied to both the training and the anomaly score. Instead of solely penalizing the reconstruction error during training, we propose adding a regularization term that penalizes the norm of the latent representation. Figure~\ref{fig:Toll} illustrates the approach.

For a training sample $x$ the loss becomes
\begin{equation}
    \mathcal{L} = \lVert\hat{x}-x\rVert_2+\beta\lVert z\rVert_2,
\end{equation}
where $z=En(x;\theta_e)$ and $\beta$ is a hyperparameter that determines the trade-off between the two terms. We modify the anomaly score accordingly to include the bottleneck representation norm:
\begin{equation}
    a(x_q) = \lVert\hat{x}_q-x_q\rVert_2+\beta\lVert z_q\rVert_2.
\end{equation}

The logic behind the adoption of this regularization is rather straightforward: due to the underlying similarity across normal samples, they should reside closely together in the latent space. A similar inductive bias is, for instance, introduced in the design of Deep Support Vector Data Description (DSVDD)~\cite{ruff2018deep}, where the encoder is tasked with learning a mapping of normal data into a hypersphere with center $c$ and radius $R$ with minimal volume, outside of which anomalies are expected to reside. Unlike in that work, we do not impose a domain constraint for the resulting distribution, but rather only push the encodings towards the origin. This addition stipulates that not only is the autoencoder forced to efficiently reconstruct the normal data, but also to map them such that they form a compact latent representation, emphasizing shared characteristics among the normal samples, thereby capturing their prevalent patterns.

Just as constraining the dimensionality of the bottleneck representation results in retaining only the most relevant information, penalizing the vector norm can serve as a selective filter for information, encouraging the encoder to identify similar attributes within the normal class, thus highlighting characteristics that are likely to differ for anomalous samples.

Not all mappings leading to small representation norms are useful in this sense. An encoder's latent space representations can be arbitrarily scaled down uniformly to reduce the norms. As another example, one can simply obtain a trivial function that maps any input to the origin. In fact,~\cite{ruff2018deep} explicitly discusses the phenomena (referred to in that work as ``hypersphere collapse'' given the constraint on latent representations) and demonstrates how it imposes restrictions on the choice of activation function and inhibits the ability to incorporate the hypersphere center and network biases as trainable parameters and instead offers an empirical strategy to pick a prespecified fixed center.

By enforcing reconstruction, the latter scenario (collapsed representation) is avoided, noting that such a representation is not an optima: a collapsed latent space results in a strictly positive reconstruction error. As such, during training the encoder is taken away from trivial collapse in the parameter space. As for the former case, in the following we will provide a theoretical analysis of the learning dynamics for a simplified case that also shows we do not merely retrieve a scaled autoencoder bottleneck representation. In fact, the learning dynamics analysis and subsequent numerical simulations demonstrate that the proposed regularization nicely complements the plain autoencoder and increases its utility for anomaly detection, which our results in Section~\ref{Results} also corroborate.

\subsection{Analysis of Learning Dynamics}

Consider matrix $X_{m\times n}$ containing the samples of a zero mean dataset, where $n$ is the number of points and $m$ is the sample dimensionality. The encoder is simplified as a single linear layer with no bias. The weights are denoted by $W_{d\times m}$ that map data to a $d-$dimensional latent space. The dataset projection can be written as
\begin{equation}
    Z=WX.
    \label{eq:z=wx}
\end{equation}
Consider the case where only the projection norm constitutes the loss. We have:
\begin{equation}
    \mathcal{L}_{norm} = \frac{1}{2n}\sum_{i,j}{Z_{ij}^2},
\end{equation}
where $Z_{ij}$ denotes the element at row $i$ and column $j$ of $Z$. We would like to see how $Z$ evolves during training. We study the dynamics via gradient flow, the continuous analog of gradient descent with vanishing step sizes (corresponding to an infinitesimal learning rate).

\begin{lemma}
    The weight matrix $W$ under gradient flow evolves by
    \begin{equation}
        \dot{W}(t)=-W(t)S,
        \label{eq:l1}
    \end{equation}
    where $S=\frac{1}{n}XX^T$ is the empirical covariance matrix.
    \label{lemma1}
\end{lemma}
The proof of Lemma~\ref{lemma1} is straightforward using the chain rule and is provided in Appendix~\ref{l1proof}.

\begin{theorem}
    The solution to the system in Eq.~\ref{eq:l1} is
    \begin{equation}
        W(t)=W(0)U\exp{(-\Lambda t)}U^T,
    \end{equation}
    where $W(0)$ are the initial weights and $\Lambda$ and $U$ are the eigenvalues and eigenvectors of $S$, respectively. As a result, the latent space evolution is given by
    \begin{equation}
        Z(t)=W(0)U\exp{(-\Lambda t)}U^TX.
    \end{equation}
    \label{theorem1}
\end{theorem}

The proof for the theorem is in Appendix~\ref{t1proof}. To interpret the evolution of latent representations, consider tentatively that $W(0)=I_m$, the $m-$dimensional identity matrix. Then we have $Z=U\exp{(-\Lambda t)}U^TX$. In the beginning, $Z(0)=X$. As time progresses, all dimensions collapse, each with an exponential shrink factor equal to the associated eigenvalue. In other words, the dimensions featuring higher variability collapse faster. Now considering a general weight matrix, we obtain a fixed projection of the shrinking dynamics, as if the input data is shrinking and being projected to a latent space through a fixed mapping.

Let us now consider the scenario in which only the reconstruction error constitutes the loss of a linear autoencoder. It is known (as shown e.g.~in~\cite{plaut2018principal}) that a linear autoencoder with a $d$-dimensional bottleneck and trained using mean squared error loss encodes data in the space spanned by the $d$ principal components (i.e.~eigenvectors associated with the $d$ largest eigenvalues of covariance), fully discarding low eigenvalue directions. Consequently, it produces a low-rank approximation of the data at reconstruction. 

\begin{figure}[!t]
    \centering
    \begin{subfigure}[b]{0.49\textwidth}
        \centering
        \includegraphics[width=1\textwidth]{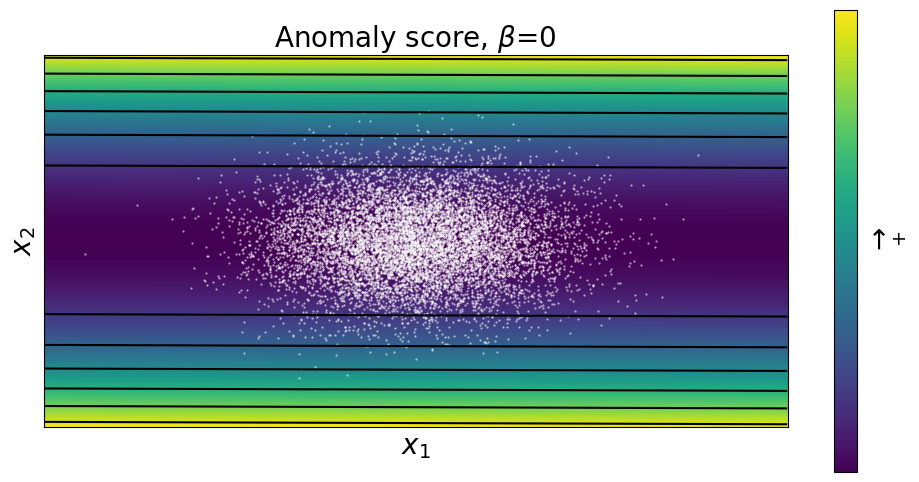}
        \caption{}
    \end{subfigure}
    \hfill
    \begin{subfigure}[b]{0.49\textwidth}
        \centering
        \includegraphics[width=1\textwidth]{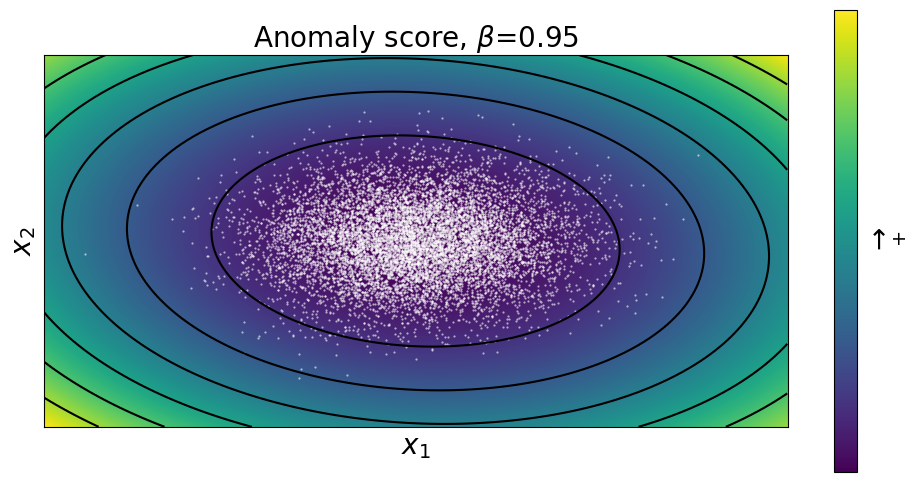}
        \caption{}
    \end{subfigure}
    \caption{Contours of anomaly score output using an unregularized (a) and a norm-regularized (b) linear autoencoder in the 2D input data space. The autoencoder is trained to encode samples from a two-dimensional Gaussian distribution (bright points) to a one-dimensional latent space and decode back to the original dimensionality. To apply regularization, a $\beta$ coefficient is multiplied by the regularization term, and $1-\beta$ is multiplied by the reconstruction error term. The unregularized autoencoder encodes the first principal component of the dataset ($x_1$) and therefore reconstruction errors scale with the distance of points from the origin along $x_2$. Regularization prevents the complete vanishing of variation along $x_2$ in the bottleneck and, for a good choice of $\beta$, the norm-regularized autoencoder yields a much more accurate anomaly score profile.}
    \label{fig:anomalyscores}
\end{figure}

Combining the effect of both loss terms, one can observe that regularizing a linear autoencoder using encoding norms counteracts the greedy elimination of information associated with dimensions featuring less pronounced variations. This can be advantageous since the dimensions featuring lower variability might in fact be important for the subsequent task (anomaly detection in this case). Figure~\ref{fig:anomalyscores} demonstrates this point. It shows contours of anomaly score when a linear autoencoder is trained on two-dimensional samples from a zero-mean bivariate Gaussian featuring a diagonal covariance with variances of 4 and 1 respectively for the first and second dimensions. Encoder and decoder are both single-layer architectures that map to and from a one-dimensional latent space. We can see that with regularization, anomaly score contours can reflect the normal class distribution much more closely. When only reconstruction is considered here, however, anomaly behavior needs to align with a strong implicit assumption for successful performance; the assumption being that the anomalous samples deviate from the distribution along the second principal component and their first principal attribute does not have a bearing on their anomalousness.

In the following (Theorem~\ref{theorem2}), we present the equations governing the learning dynamics for a linear autoencoder whose training is regularized via encoding norm. Encoder and decoder are linear layers with no bias whose weights are respectively $W_1$ and $W_2$. $\hat{X}=W_2W_1X$ is the reconstructed dataset. The loss function is a weighted sum of MSE reconstruction loss $\mathcal{L}_{rec}$ and $\mathcal{L}_{norm}$:
\begin{equation}
    \mathcal{L} = \left[\frac{1}{2n}\sum_{i,j}{(\hat{X}_{ij}-X_{ij})^2}\right]+\beta\left[\frac{1}{2n}\sum_{i,j}{Z_{ij}^2}\right].
\end{equation}

\begin{theorem}
    $W_1$ and $W_2$ under gradient flow evolve by
    \begin{subequations}
    \begin{align}
        \dot{W}_1(t)&=[W_2^T-W_2^TW_2W_1-\beta W_1]S,\\
        \dot{W}_2(t)&=[I-W_2W_1]SW_1^T.
    \end{align}
    \end{subequations}
    \label{theorem2}
\end{theorem}

Proof of Theorem~\ref{theorem2} can be found in Appendix~\ref{t2proof}. Even in such a simplified case, the equations are nonlinear and coupled, making further analysis challenging. Hence, these equations remain unsolved in this work.

While the above analysis does not consider nonlinearities, it provides insight into the behavior of an anomaly detection scheme built based on autoencoding performance. Based on our observation, by properly selecting $\beta$, suitable expressions of data manifold can be retrieved for anomaly detection, with virtually no computational or memory overhead during training and inference. Also given the technique and the principle behind it, the algorithm can readily accommodate various data modalities.

\section{Experiments}\label{Results}

\subsection{Datasets and Implementation Details}

\begin{figure}[!t]
    \centering
    \begin{subfigure}[b]{0.49\textwidth}
        \centering
        \includegraphics[width=1\textwidth]{mnistfmnistcifar100.png}
        \caption{}
    \end{subfigure}
    \hfill
    \begin{subfigure}[b]{0.49\textwidth}
        \centering
        \includegraphics[width=1\textwidth]{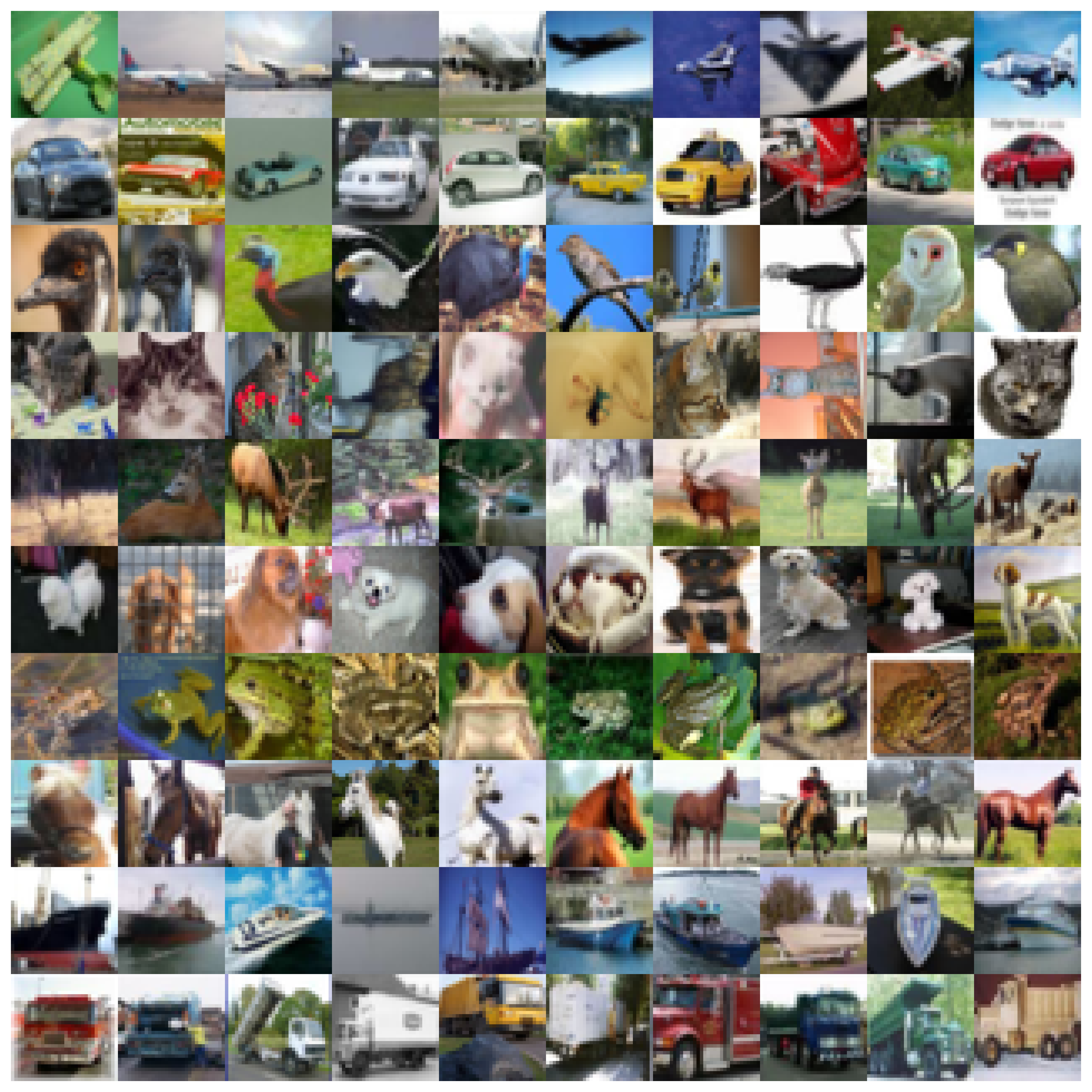}
        \caption{}
    \end{subfigure}
    \caption{Samples from MNIST (a, top left), Fashion-MNIST (a, bottom left), CIFAR-100 (a, right), and CIFAR-10 (b) datasets. Rows show different classes.}
    \label{fig:MNIST_CIFAR}
\end{figure}

We present the results of Toll on five datasets: MNIST~\cite{lecun2010mnist}, Fashion-MNIST~\cite{xiao2017fashion}, CIFAR-10~\cite{krizhevsky2009learning}, CIFAR-100~\cite{krizhevsky2009learning}, and Arrhythmia~\cite{dua2017uci}. For MNIST, Fashion-MNIST, CIFAR-10, and CIFAR-100 we use the same protocol as that used in e.g.~\cite{ruff2018deep,perera2019ocgan,yang2021memory}. MNIST, Fashion-MNIST, CIFAR-10, and CIFAR-100 are visual datasets. MNIST, Fashion-MNIST, and CIFAR-10 all have 10 classes. CIFAR-100 consists of 20 superclasses used in this task, each containing 5 closely related classes. For instance, the superclass \emph{aquatic mammals} contains classes \emph{beaver, dolphin, otter, seal}, and \emph{whale}. These datasets respectively have 60000, 60000, 50000, and 50000 images in their training set and 10000 images in test sets. Figure~\ref{fig:MNIST_CIFAR} shows sample images from all classes in the four visual datasets. We also set aside 10000 samples from the training set for validation. In all datasets, we perform 10 experiments, each time considering one of the classes as the normal class and all the other classes as anomalous. Training is done on samples from the normal class of the associated experiment. (i.e.~approximately 5000 samples for MNIST and Fashion-MNIST, 4000 for CIFAR-10, and 2000 for CIFAR-100).

The Arrhythmia dataset is a tabular dataset adapted from a dataset for the classification of cardiac arrhythmia. The dataset consists of a total of 452 samples and 274 attributes across 16 classes. In line with previous work, we designate classes 3, 4, 5, 7, 8, 9, 14, and 15 as anomalous and the remaining classes as normal. The anomalous samples make up approximately 15 percent of the dataset. We set aside 20 percent of the data for testing and consider another 30 percent for validation and use the remaining normal samples to train the model.

MNIST and Fashion-MNIST datasets were normalized to have a zero mean and a standard deviation of 1 for pixel values. CIFAR-10 images were preprocessed with global contrast normalization using the $L^1$-norm and rescaled to $\left[ 0,1\right]$ with min-max scaling. The encoder and decoder architectures used for CIFAR-10 and CIFAR-100 experiments are similar to encoder and generator architectures in~\cite{dumoulin2016adversarially,yang2021memory}. The architectures for MNIST, Fashion-MNIST, and Arrhythmia experiments are also based on~\cite{yang2021memory}. See Appendix~\ref{arch} for detailed architectures.

In all experiments, training is done through iterating over 2000 random batches with checkpoints considered at 20 batch intervals. The batch sizes for MNIST, Fashion-MNIST, CIFAR-10, CIFAR-100, and Arrhythmia are respectively 1000, 500, 50, 100, and 100. Adam optimizer with a fixed learning rate is used for training. Learning rates, $\beta$ values, and the bottleneck dimensions are picked based on the validation set performance. We keep these hyperparameters unchanged across all normal classes in the four visual datasets, but we note that test performances further improve, in many instances considerably, if hyperparameters are picked specific to the normal class. The best-performing model on the validation set is selected for subsequent evaluation on the test set. Each experiment is repeated for 10 different random seeds and averaged results are reported.

We comply with the existing literature for evaluation. The metric used to evaluate results on MNIST, Fashion-MNIST, CIFAR-10, and CIFAR-100 is the area under ROC curve (AUC). For Arrhythmia, we set the score threshold such that we label the expected 15 percent of samples with the highest anomaly scores as anomalous. We then calculate the $F_1$ score.

To further demonstrate the utility of targeted collapse, we also apply the collapse idea in the context of a well-performing anomaly detection method~\cite{mirzaei2022fake}. In this technique, which we refer to as FITYMI (Fake It Till You Make It) following the paper title, the authors first generate synthetic anomalous data using a diffusion model~\cite{song2020score} and then fine-tune a binary classifier that incorporates a strong pre-trained backbone. The backbone mainly considered in that work is a pre-trained vision transformer~\cite{dosovitskiy2020image}. Specifically, ViT-B/16 pre-trained on ImageNet-21K~\cite{deng2009imagenet} is adopted.

We replicate the FITYMI experiments on CIFAR-10 and CIFAR-100 datasets in the exact manner described in the paper using the code and generated anomalous samples provided by the authors with no modifications. To then test the targeted collapse idea, we simply add the collapse term for normal samples to the classification loss during training without any other modifications of the parameters and the pipeline. Given the pre-trained backbone in this case, the target for collapse in the latent space was set as the mean of the normal samples, as approximated per batch, in order to simplify the implementation. We again repeat each experiment 10 times and report averaged results.

\subsection{Results}

Tables~\ref{tab:mnist},~\ref{tab:cifar-10},~\ref{tab:fmnist}, and~\ref{tab:arrhythmia} show anomaly detection results on the datasets. On all datasets, our approach achieves a competitive performance and outperforms several well-known and considerably more sophisticated methods in various normal class experiments and on average for some datasets. Notably, GeoTrans~\cite{golan2018deep} and ARNet~\cite{huang2019attribute} incorporate significantly larger, more sophisticated network architectures, as well as data augmentation and self-supervised training objectives. As the tables show, norm regularization is a simple yet powerful way to improve anomaly detection performance.

Tables~\ref{tab:fitymi_cifar10} and \ref{tab:fitymi_cifar100} demonstrate the effect of targeted collapse on the performance of FITYMI as a strong anomaly detection technique based on using powerful pre-trained backbones and synthetic out-of-distribution data. Even in this high-performance regime, the addition of targeted collapse is capable of further improving the model performance virtually at no cost and despite the availability of out-of-distribution data that can already largely inform the model of the desirable separation in the latent space.

\begin{table*}
\caption{Mean and standard deviation of AUC in \% on MNIST classes over 10 seeds. The highest performance in each row is in bold.}
\centering
\tiny
\centerline{
\begin{tabular}{c @{\hspace{6pt}}|c @{\hspace{6pt}}c @{\hspace{6pt}}c @{\hspace{6pt}}c @{\hspace{6pt}}c @{\hspace{6pt}}c @{\hspace{6pt}}c @{\hspace{6pt}}c @{\hspace{6pt}}c @{\hspace{6pt}}c @{\hspace{6pt}}c @{\hspace{6pt}}}
\toprule
Normal & OC-SVM~\cite{scholkopf1999support} & OCGAN~\cite{perera2019ocgan} & VAE~\cite{kingma2013auto} & DCAE~\cite{makhzani2015winner} & IF~\cite{liu2008isolation} & AnoGAN~\cite{schlegl2017unsupervised} & KDE~\cite{parzen1962estimation} & DSVDD~\cite{ruff2018deep} & MEMAE~\cite{gong2019memorizing} & MEMGAN~\cite{yang2021memory} & \textbf{Toll}\\
\midrule
0 & 98.6$\pm$0.0 & \textbf{99.8} & 99.7 & 97.6$\pm$0.7 & 98.0$\pm$0.3 & 96.6$\pm$1.3 & 97.1$\pm$0.0 & 98.0$\pm$0.7 & 99.3$\pm$0.1 & 99.3$\pm$0.1 & \textbf{99.8$\pm$0.1}\\
1 & 99.5$\pm$0.0 & \textbf{99.9} & \textbf{99.9} & 98.3$\pm$0.6 & 97.3$\pm$0.4 & 99.2$\pm$0.6 & 98.9$\pm$0.0 & 99.7$\pm$0.1 & 99.8$\pm$0.0 & \textbf{99.9$\pm$0.0} & \textbf{99.9$\pm$0.0}\\
2 & 82.5$\pm$0.1 & 94.2 & 93.6 & 85.4$\pm$2.4 & 88.6$\pm$0.5 & 85.0$\pm$2.9 & 79.0$\pm$0.0 & 91.7$\pm$0.8 & 90.6$\pm$0.8& 94.5$\pm$0.1 & \textbf{96.2$\pm$1.6}\\
3 & 88.1$\pm$0.0 & 96.3 & 95.9 & 86.7$\pm$0.9 & 89.9$\pm$0.4 & 88.7$\pm$2.1 & 86.2$\pm$0.0 & 91.9$\pm$1.5 & 94.7$\pm$0.6 & 95.7$\pm$0.4 & \textbf{97.9$\pm$0.6}\\
4 & 94.9$\pm$0.0 & 97.5 & 97.3 & 86.5$\pm$2.0 & 92.7$\pm$0.6 & 89.4$\pm$1.3 & 87.9$\pm$0.0 & 94.9$\pm$0.8 & 94.5$\pm$0.4 & 96.1$\pm$0.4 & \textbf{97.8$\pm$0.4}\\
5 & 77.1$\pm$0.0 & 98.0 & 96.4 & 78.2$\pm$2.7 & 85.5$\pm$0.8 & 88.3$\pm$2.9 & 73.8$\pm$0.0 & 88.5$\pm$0.9 & 95.1$\pm$0.1 & 93.6$\pm$0.3 & \textbf{98.1$\pm$0.6}\\
6 & 96.5$\pm$0.0 & 99.1 & 99.3 & 94.6$\pm$0.5 & 95.6$\pm$0.3 & 94.7$\pm$2.7 & 87.6$\pm$0.0 & 98.3$\pm$0.5 & 98.4$\pm$0.5 & 98.6$\pm$0.1 & \textbf{99.5$\pm$0.1}\\
7 & 93.7$\pm$0.0 & 98.1 & 97.6 & 92.3$\pm$1.0 & 92.0$\pm$0.4 & 93.5$\pm$1.8 & 91.4$\pm$0.0 & 94.6$\pm$0.9 & 95.4$\pm$0.2 & 96.2$\pm$0.2 & \textbf{98.7$\pm$0.2}\\
8 & 88.9$\pm$0.0 & 93.9 & 92.3 & 86.5$\pm$1.6 & 89.9$\pm$0.4 & 84.9$\pm$2.1 & 79.2$\pm$0.0 & 93.9$\pm$1.6 & 86.9$\pm$0.5 & 93.5$\pm$0.1 & \textbf{97.3$\pm$0.5}\\
9 & 93.1$\pm$0.0 & 98.1 & 97.6 & 90.4$\pm$1.8 & 93.5$\pm$0.3 & 92.4$\pm$1.1 & 88.2$\pm$0.0 & 96.5$\pm$0.3 & 97.3$\pm$0.2 & 95.9$\pm$0.1 & \textbf{98.5$\pm$0.3}\\
\midrule
Average & 91.3 & 97.5 & 97.0 & 89.7 & 92.3 & 91.4 & 87.0 & 94.8 & 95.2 & 96.5 & \textbf{98.4}\\
\bottomrule
\end{tabular}}
\label{tab:mnist}
\end{table*}
\begin{table*}
\caption{Mean and standard deviation of AUC in \% on CIFAR-10 classes over 10 seeds. The highest performance in each row is in bold.}
\centering
\tiny
\centerline{
\begin{tabular}{r|ccccccccccc}
\toprule
Normal & OCGAN & VAE & DSVDD & DSEBM~\cite{zhai2016deep} & DAGMM~\cite{zong2018deep} & IF & AnoGAN & ALAD~\cite{zenati2018adversarially} & MEMAE & MEMGAN & \textbf{Toll}\\
\midrule
airplane & \textbf{75.7} & 70.0 & 61.7$\pm$4.1 & 41.4$\pm$2.3 & 56.0$\pm$6.9 & 60.1$\pm$0.7 & 67.1$\pm$2.5 & 64.7$\pm$2.6 & 66.5$\pm$0.9 & 73.0$\pm$0.8 & 75.0$\pm$4.3\\
auto. & 53.1 & 38.6 & 65.9$\pm$2.1 & 57.1$\pm$2.0 & 48.3$\pm$1.8 & 50.8$\pm$0.6 & 54.7$\pm$3.4 & 38.7$\pm$0.8 & 46.4$\pm$0.1 & 52.5$\pm$0.7 & \textbf{67.6$\pm$3.2}\\
bird & 64.0 & \textbf{67.9} & 50.8$\pm$0.8 & 61.9$\pm$0.1 & 53.8$\pm$4.0 & 49.2$\pm$0.4 & 52.9$\pm$3.0 & 67.0$\pm$0.7 & 66.0$\pm$0.1 & 67.2$\pm$0.1 & 56.2$\pm$2.8\\
cat & 62.0 & 53.5 & 59.1$\pm$1.4 & 50.1$\pm$0.4 & 51.2$\pm$0.8 & 55.1$\pm$0.4 & 54.5$\pm$1.9 & 59.2$\pm$0.3 & 52.9$\pm$0.1 & 57.3$\pm$0.2 & \textbf{64.6$\pm$1.6}\\
deer & 72.3     & \textbf{74.8} & 60.9$\pm$1.1 & 73.3$\pm$0.2 & 52.2$\pm$7.3 & 49.8$\pm$0.4 & 65.1$\pm$3.2 & 72.7$\pm$0.6 & 72.8$\pm$0.1 & 73.9$\pm$0.9 & 67.7$\pm$1.4\\
dog & 62.0      & 52.3 & 65.7$\pm$0.8 & 60.5$\pm$0.3 & 49.3$\pm$3.6 & 58.5$\pm$0.4 & 60.3$\pm$2.6 & 52.8$\pm$1.2 & 52.9$\pm$0.2 & 65.0$\pm$0.2 & \textbf{68.1$\pm$1.6}\\
frog & 72.3     & 68.7 & 67.7$\pm$2.6 & 68.4$\pm$0.3 & 64.9$\pm$1.7 & 42.9$\pm$0.6 & 58.5$\pm$1.4 & 69.5$\pm$1.1 & 63.7$\pm$0.4 & 72.8$\pm$0.7 & \textbf{74.7$\pm$2.5}\\
horse & 57.5    & 49.3 & \textbf{67.3$\pm$0.9} & 53.3$\pm$0.7 & 55.3$\pm$0.8 & 55.1$\pm$0.7 & 62.5$\pm$0.8 & 44.8$\pm$0.4 & 45.9$\pm$0.1 & 52.5$\pm$0.5 & 65.8$\pm$1.3\\
ship & \textbf{82.0}     & 69.6 & 75.9$\pm$1.2 & 73.9$\pm$0.3 & 51.9$\pm$2.4 & 74.2$\pm$0.6 & 75.8$\pm$4.1 & 73.4$\pm$0.4 & 70.1$\pm$0.1 & 74.4$\pm$0.3 & 80.0$\pm$1.1\\
truck & 55.4    & 38.6 & 73.1$\pm$1.2 & 63.6$\pm$3.1 & 54.2$\pm$5.8 & 58.9$\pm$0.7 & 66.5$\pm$2.8 & 39.2$\pm$1.3 & 48.2$\pm$0.2 & 65.6$\pm$1.6 & \textbf{76.2$\pm$0.5}\\
\midrule
Average & 65.6     & 58.3 & 64.8 & 60.4 & 54.4 & 55.5 & 61.8 &  59.3 & 58.5 & 65.3 & \textbf{69.6}\\
\bottomrule
\end{tabular}}
\label{tab:cifar-10}
\end{table*}

\begin{table*}
\caption{Mean and standard deviation of AUC in \% on Fashion-MNIST and CIFAR-100 classes over 10 seeds. The highest performance in each row is in bold.}
\centering
\tiny
\centerline{
\begin{tabular}{lr|ccccccc}
\toprule
Dataset & Normal & DAGMM & DSEBM & ADGAN~\cite{deecke2018anomaly} & GANomaly~\cite{akcay2019ganomaly} & GeoTrans~\cite{golan2018deep} & ARNet~\cite{huang2019attribute} & \textbf{Toll}\\
\midrule
 & T-shirt/top & 42.1 & 91.6 & 89.9 & 80.3 & \textbf{99.4} & 92.7 & 91.9$\pm$0.2 \\
 & Trouser & 55.1 & 71.8 & 81.9 & 83.0 & 97.6 & \textbf{99.3} & 98.7$\pm$0.0 \\
 & Pullover & 50.4 & 88.3 & 87.6 & 75.9 & 91.1 & 89.1 & \textbf{91.3$\pm$0.2} \\
 & Dress & 57.0 & 87.3 & 91.2 & 87.2 & 89.9 & \textbf{93.6} & 92.4$\pm$0.2 \\
\multirow{2}{6em}{Fashion-MNIST} & Coat & 26.9 & 85.2 & 86.5 & 71.4 & \textbf{92.1} & 90.8 & 91.5$\pm$0.1 \\
 & Sandal & 70.5 & 87.1 & 89.6 & 92.7 & \textbf{93.4} & 93.1 & 93.1$\pm$0.2 \\
 & Shirt & 48.3 & 73.4 & 74.3 & 81.0 & 83.3 & \textbf{85.0} & 84.6$\pm$0.3 \\
 & Sneaker & 83.5 & 98.1 & 97.2 & 88.3 & 98.9 & 98.4 & \textbf{99.0$\pm$0.0} \\
 & Bag & 49.9 & 86.0 & 89.0 & 69.3 & 90.8 & \textbf{97.8} & 90.8$\pm$0.6 \\
 & Ankle boot & 34.0 & 97.1 & 97.1 & 80.3 & \textbf{99.2} & 98.4 & 98.9$\pm$0.1 \\
\cmidrule{2-9}
 & Average & 51.8 & 86.6 & 88.4 & 80.9 & 93.5 & \textbf{93.9} & 93.2 \\
\midrule
\midrule
 & aquatic mammals & 43.4 & 64.0 & 63.1 & 57.9 & 74.7 & \textbf{77.5} & 70.3$\pm$3.4 \\
 & fish & 49.5 & 47.9 & 54.9 & 51.9 & 68.5 & \textbf{70.0} & 66.2$\pm$1.1 \\
 & flowers & 66.1 & 53.7 & 41.3 & 36.0 & 74.0 & 62.4 & \textbf{80.3$\pm$0.7} \\
 & food containers & 52.6 & 48.4 & 50.0 & 46.5 & \textbf{81.0} & 76.2 & 58.2$\pm$1.5 \\
 & fruit and vegetables & 56.9 & 59.7 & 40.6 & 46.6 & 78.4 & 77.7 & \textbf{80.1$\pm$1.3} \\
 & household electrical devices & 52.4 & 46.6 & 42.8 & 42.9 & 59.1 & \textbf{64.0} &61.4$\pm$2.6 \\
 & household furniture & 55.0 & 51.7 & 51.1 & 53.7 & 81.8 & \textbf{86.9} & 64.8$\pm$3.0 \\
 & insects & 52.8 & 54.8 & 55.4 & 59.4 & 65.0 & 65.6 & \textbf{70.5$\pm$1.3} \\
 & large carnivores & 53.2 & 66.7 & 59.2 & 63.7 & \textbf{85.5} & 82.7 & 62.1$\pm$2.1 \\
\multirow{2}{6em}{CIFAR-100} & large man-made outdoor things & 42.5 & 71.2 & 62.7 & 68.0 & \textbf{90.6} & 90.2 & 79.6$\pm$1.4 \\
 & large natural outdoor scenes & 52.7 & 78.3 & 79.8 & 75.6 & \textbf{87.6} & 85.9 & 81.6$\pm$1.1 \\
 & large omnivores and herbivores & 46.4 & 62.7 & 53.7 & 57.6 & \textbf{83.9} & 83.5 & 62.6$\pm$1.2 \\
 & medium-sized mammals & 42.7 & 66.8 & 58.9 & 58.7 & 83.2 & \textbf{84.6} & 61.2$\pm$1.9 \\
 & non-insect invertebrates & 45.4 & 52.6 & 57.4 & 59.9 & 58.0 & \textbf{67.6} & 58.6$\pm$2.0 \\
 & people & 57.2 & 44.0 & 39.4 & 43.9 & \textbf{92.1} & 84.2 & 65.4$\pm$4.5 \\
 & reptiles & 48.8 & 56.8 & 55.6 & 59.9 & 68.3 & \textbf{74.1} & 56.1$\pm$1.5 \\
 & small mammals & 54.4 & 63.1 & 63.3 & 64.4 & 73.5 & \textbf{80.3} & 62.3$\pm$1.1 \\
 & trees & 36.4 & 73.0 & 66.7 & 71.8 & \textbf{93.8} & 91.0 & 80.9$\pm$1.8 \\
 & vehicles 1 & 52.4 & 57.7 & 44.3 & 54.9 & \textbf{90.7} & 85.3 & 65.2$\pm$1.4 \\
 & vehicles 2 & 50.3 & 55.5 & 53.0 & 56.8 & 85.0 & \textbf{85.4} & 63.6$\pm$1.6 \\
\cmidrule{2-9}
 & Average & 50.5 & 58.8 & 54.7 & 56.5 & 78.7 & \textbf{78.8} & 67.6 \\
 \bottomrule
\end{tabular}}
\label{tab:fmnist}
\end{table*}
\begin{table*}
\caption{Average F$_{1}$ score in \% on Arrhythmia dataset over 10 seeds. The highest performance is in bold.}
\centering
\tiny
\centerline{
\begin{tabular}{cccccccccccc}
\toprule
IF & OC-SVM & DSEBM-e & AnoGAN & DAGMM & ALAD & RCGAN~\cite{yang2020regularized} & DSVDD & GOAD~\cite{bergman2020classification} & MEMAE & MEMGAN & \textbf{Toll}\\
\midrule
53.03 & 45.18 & 46.01 & 42.42 & 49.83 & 51.52 & 54.14 & 34.79 & 52.00 & 51.13 & 55.72 & \textbf{60.00}\\
\bottomrule
\end{tabular}}
\label{tab:arrhythmia}
\end{table*}
\begin{table*}
\caption{The effect of targeted collapse on the performance of FITYMI evaluated on CIFAR-10. Reported results are the mean and standard deviation of AUC in \% for each class over 10 seeds. The highest performance in each experiment is in bold.}
\centering
\tiny
\centerline{
\begin{tabular}{l @{\hspace{6pt}}|c @{\hspace{6pt}}c @{\hspace{6pt}}c @{\hspace{6pt}}c @{\hspace{6pt}}c @{\hspace{6pt}}c @{\hspace{6pt}}c @{\hspace{6pt}}c @{\hspace{6pt}}c @{\hspace{6pt}}c @{\hspace{6pt}}|c @{\hspace{6pt}}}
\toprule
Normal & 0 & 1 & 2 & 3 & 4 & 5 & 6 & 7 & 8 & 9 & Average \\
\midrule
FITYMI~\cite{mirzaei2022fake} & \textbf{99.25$\pm$0.07} & 98.63$\pm$0.26 & 97.08$\pm$0.94 & 98.10$\pm$0.17 & 98.93$\pm$0.27 & 96.06$\pm$0.56 & 99.36$\pm$0.10 & 99.53$\pm$0.05 & 98.88$\pm$0.55 & 98.29$\pm$1.23 & 98.41$\pm$0.19 \\
FITYMI+Toll & 99.17$\pm$0.05 & \textbf{99.23$\pm$0.21} & \textbf{97.42$\pm$0.38} & \textbf{98.14$\pm$0.17} & \textbf{99.23$\pm$0.10} & \textbf{97.25$\pm$0.42} & \textbf{99.60$\pm$0.05} & \textbf{99.64$\pm$0.06} & \textbf{99.48$\pm$0.14} & \textbf{98.62$\pm$0.17} & \textbf{98.78$\pm$0.07} \\
\bottomrule
\end{tabular}}
\label{tab:fitymi_cifar10}
\end{table*}
\begin{table*}
\caption{The effect of targeted collapse on the performance of FITYMI evaluated on CIFAR-100. Reported results are the mean and standard deviation of AUC in \% for each class over 10 seeds. The highest performance in each experiment is in bold.}
\centering
\tiny
\centerline{
\begin{tabular}{l @{\hspace{6pt}}|c @{\hspace{6pt}}c @{\hspace{6pt}}c @{\hspace{6pt}}c @{\hspace{6pt}}c @{\hspace{6pt}}c @{\hspace{6pt}}c @{\hspace{6pt}}c @{\hspace{6pt}}c @{\hspace{6pt}}c @{\hspace{6pt}}c @{\hspace{6pt}}}
\toprule
Normal & 0 & 1 & 2 & 3 & 4 & 5 & 6 & 7 & 8 & 9 & 10 \\
\midrule
FITYMI~\cite{mirzaei2022fake} & 95.41$\pm$0.30 & 95.44$\pm$0.80 & 97.56$\pm$0.75 & 96.02$\pm$1.05 & 96.51$\pm$0.34 & 96.09$\pm$0.67 & 98.42$\pm$1.09 & 96.02$\pm$0.34 & 97.36$\pm$0.21 & \textbf{96.93$\pm$0.32} & \textbf{96.44$\pm$0.51} \\
FITYMI+Toll & \textbf{97.29$\pm$0.32} & \textbf{96.21$\pm$0.94} & \textbf{97.57$\pm$1.67} & \textbf{96.30$\pm$1.50} & \textbf{97.33$\pm$0.22} & \textbf{97.05$\pm$0.44} & \textbf{99.02$\pm$0.06} & \textbf{96.83$\pm$0.51} & \textbf{98.43$\pm$0.11} & 96.84$\pm$0.65 & 95.86$\pm$0.70 \\
\midrule
\midrule
Normal & 11 & 12 & 13 & 14 & 15 & 16 & 17 & 18 & 19 & \multicolumn{2}{|c}{Average} \\
\midrule
FITYMI~\cite{mirzaei2022fake} & 96.91$\pm$0.21 & 96.89$\pm$0.20 & 93.43$\pm$0.62 & 96.93$\pm$1.41 & 95.12$\pm$0.50 & 96.88$\pm$0.25 & 96.25$\pm$0.73 & 98.46$\pm$0.29 & 96.12$\pm$0.44 & \multicolumn{2}{|c}{96.46$\pm$0.11} \\ 
FITYMI+Toll & \textbf{97.73$\pm$0.12} & \textbf{97.80$\pm$0.17} & \textbf{93.99$\pm$0.94} & \textbf{97.18$\pm$0.95} & \textbf{96.69$\pm$0.32} & \textbf{97.53$\pm$0.15} & \textbf{96.82$\pm$1.11} & \textbf{98.74$\pm$0.16} & \textbf{97.04$\pm$0.29} & \multicolumn{2}{|c}{\textbf{97.11$\pm$0.14}} \\
\bottomrule
\end{tabular}}
\label{tab:fitymi_cifar100}
\end{table*}

\subsection{Ablation Study}

\begin{figure}[!t]
\centering
  \includegraphics[width=0.6\linewidth]{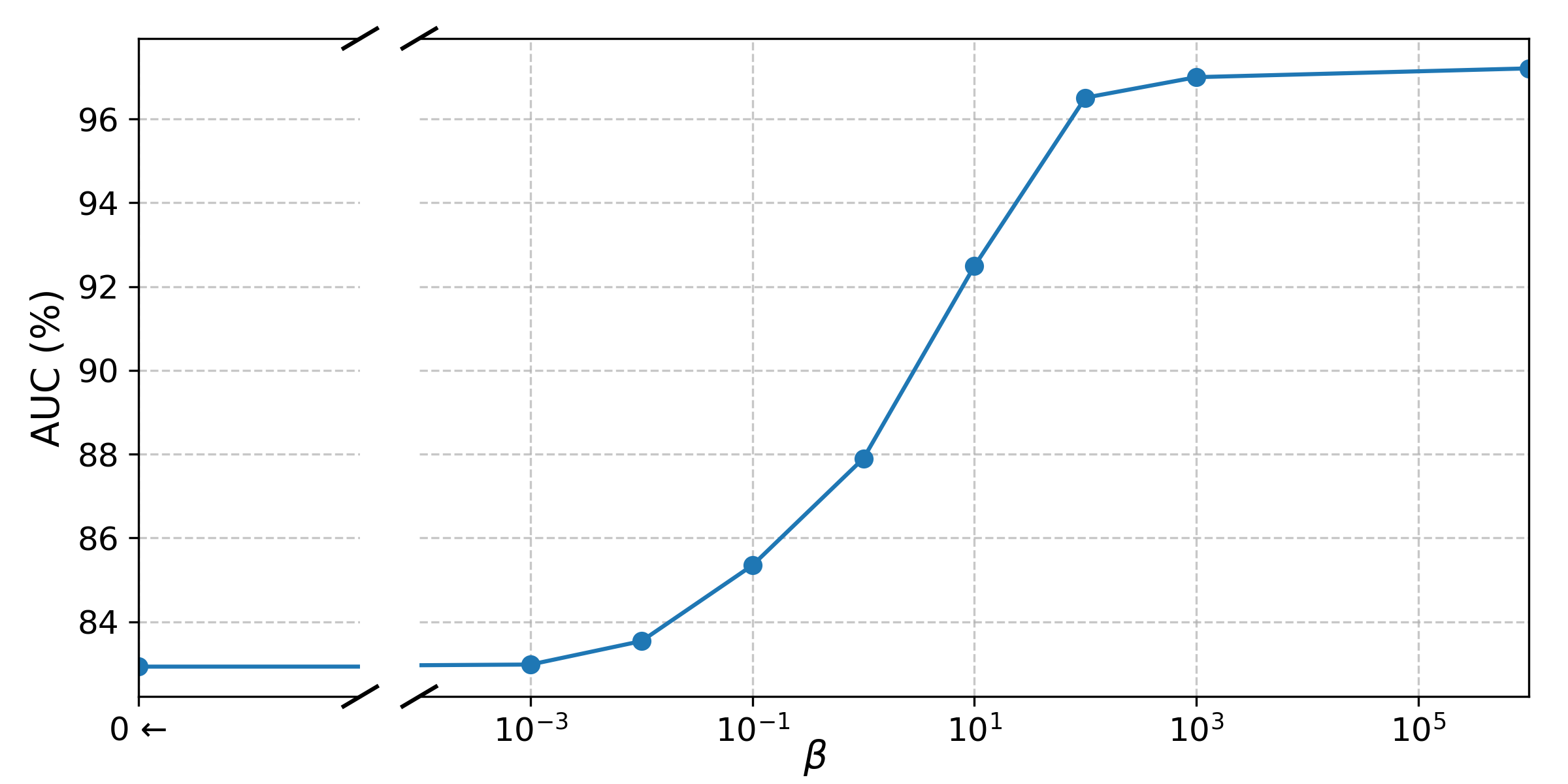}
\caption{Effect of regularization intensity for digit ``8'' as normal class in MNIST. The AUC values shown are averaged over 3 random seeds.}
\label{fig:performancevsbeta}
\end{figure}

Table~\ref{tab:beta8} compares the anomaly detection performance with and without norm minimization. The effect of regularization varies across different normal classes. This is due to the fact that different classes benefit to different extents from the regularization and that the single $\beta$ value used across various classes is not necessarily a good fit for all. Even with this caveat, the contribution of norm minimization is significant at about 4 percent on average in this high-accuracy regime. For the given $\beta$, this effect is the most pronounced when considering digit ``8'' as the normal class. Figure~\ref{fig:performancevsbeta} shows the effect of introducing the regularization on performance at various intensities.
\begin{table*}
\caption{AUC comparison for MNIST classes over 10 seeds for an unregularized versus a regularized autoencoder. The highest performance in each row is in bold.}
\centering
\tiny
\centerline{
\begin{tabular}{c|cc}
\toprule
Normal & $\beta$=0 & $\beta$=1000\\
\midrule
0 & 99.1$\pm$0.1 & \textbf{99.8$\pm$0.1}\\
1 & \textbf{99.9$\pm$0.0} & \textbf{99.9$\pm$0.0}\\
2 & 91.4$\pm$0.3 & \textbf{96.2$\pm$1.6}\\
3 & 93.4$\pm$0.2 & \textbf{97.9$\pm$0.6}\\
4 & 94.3$\pm$0.3 & \textbf{97.8$\pm$0.4}\\
5 & 95.1$\pm$0.1 & \textbf{98.1$\pm$0.6}\\
6 & 98.5$\pm$0.1 & \textbf{99.5$\pm$0.1}\\
7 & 96.4$\pm$0.1 & \textbf{98.7$\pm$0.2}\\
8 & 82.9$\pm$0.2 & \textbf{97.3$\pm$0.5}\\
9 & 96.2$\pm$0.1 & \textbf{98.5$\pm$0.3}\\
\midrule
Average & 94.7 & \textbf{98.4}\\
\bottomrule
\end{tabular}}
\label{tab:beta8}
\end{table*}

In our method, we have incorporated the Euclidean norm. In principle, other norms such as $L_1$ can also be adopted. Table~\ref{tab:norm} demonstrates the effect of the choice of norm for the MNIST dataset. For the $L_1$ and $L_2$ norms, we obtain a similar level of performance, while performance is lowered with $L_\infty$.
\begin{table*}
\caption{AUC comparison for MNIST classes over 10 seeds for using $L_1$, $L_2$, and $L_\infty$ norms of the bottleneck representations. The highest performance in each row is in bold.}
\centering
\tiny
\centerline{
\begin{tabular}{c|ccc}
\toprule
Normal & $1$-norm & $2$-norm & $\infty$-norm\\
\midrule
0 & 99.7$\pm$0.1 & \textbf{99.8$\pm$0.1} & 99.3$\pm$0.3\\
1 & 99.8$\pm$0.0 & \textbf{99.9$\pm$0.0} & 99.7$\pm$0.1\\
2 & \textbf{96.4$\pm$1.6} & 96.2$\pm$1.6 & 94.4$\pm$2.0\\
3 & 97.8$\pm$0.8 & \textbf{97.9$\pm$0.6} & 97.3$\pm$0.4\\
4 & \textbf{97.8$\pm$0.3} & \textbf{97.8$\pm$0.4} & 96.8$\pm$0.2\\
5 & \textbf{98.3$\pm$0.6} & 98.1$\pm$0.6 & 96.9$\pm$0.8\\
6 & \textbf{99.5$\pm$0.1} & \textbf{99.5$\pm$0.1} & 99.1$\pm$0.3\\
7 & 98.6$\pm$0.3 & \textbf{98.7$\pm$0.2} & 97.6$\pm$0.3\\
8 & 97.1$\pm$0.5 & \textbf{97.3$\pm$0.5} & 95.2$\pm$0.8\\
9 & \textbf{98.5$\pm$0.3} & \textbf{98.5$\pm$0.3} & 97.3$\pm$0.3\\
\midrule
Average & \textbf{98.4} & \textbf{98.4} & 97.4\\
\bottomrule
\end{tabular}}
\label{tab:norm}
\end{table*}

\section{Conclusion}\label{Conclusion}
In this work, we proposed a simple method for regularizing autoencoder training and modifying the anomaly score function based on minimizing the norm of bottleneck representations. The method comes from simple first principles and despite its simplicity, appropriately fits the purpose of anomaly detection, leading to a significant performance boost. The implementation is straightforward and the methodology does not introduce any constraints on the application or the associated data modality. We showcased the technique's advantages by conducting comparisons of our method against well-established and intricate approaches to anomaly detection, as well as through ablation studies that evaluate the impact of the regularization effect. We further demonstrated that the idea of targeted collapse can sit well with cutting-edge techniques and can still improve their performance despite their extensively pre-trained foundation models and access to anomalous samples during training. We also carried out an analytical investigation into the learning dynamics, exploring the phenomenological consequences of the proposed modification and how it complements the capabilities of autoencoders.

\section*{Acknowledgments}
This work was supported by Ford Motor Company.

\FloatBarrier

\bibliographystyle{plain}
\bibliography{sample}

\newpage
\appendix
\section*{Appendix}

\section{Delayed Proofs}
\subsection{Proof of Lemma~\ref{lemma1}}{\label{l1proof}}

The governing equation for weight dynamics in gradient flow is

\begin{equation}
    \dot{W}=-\frac{\partial\mathcal{L}_{norm}}{\partial W}.
\end{equation}

Using chain rule we can write

\begin{align}
    \frac{\partial\mathcal{L}_{norm}}{\partial W}&=\frac{\partial\mathcal{L}_{norm}}{\partial Z}\frac{\partial Z}{\partial W}\notag\\\notag\\
    &=\frac{\partial\mathcal{L}_{norm}}{\partial Z}X^T\notag\\\notag\\
    &=\frac{1}{n}ZX^T\notag\\\notag\\
    &=\frac{1}{n}WXX^T\notag\\\notag\\
    &=WS;
\end{align}

therefore,

\begin{equation}
        \dot{W}(t)=-W(t)S.\notag
\end{equation}
\subsection{Proof of Theorem~\ref{theorem1}}{\label{t1proof}}

The solution to the system in Eq.~\ref{eq:l1} can be expressed as

\begin{equation}
        W(t)=W(0)\exp{(-St)}.
\end{equation}

If we write the eigendecomposition of the empirical covariance as $S=U\Lambda U^T$ (considering the orthonormality of $U$) and substitute above we get

\begin{equation}
    W(t)=W(0)\exp{(-U\Lambda U^Tt)},
\end{equation}

which we can further evaluate as 

\begin{equation}
    W(t)=W(0)U\exp{(-\Lambda t)}U^T.\notag
\end{equation}

Using Eq.~\ref{eq:z=wx} we can subsequently evaluate representation dynamics:

\begin{equation}
    Z(t)=W(0)U\exp{(-\Lambda t)}U^TX.\notag
\end{equation}

\subsection{Proof of Theorem~\ref{theorem2}}{\label{t2proof}}

The governing equations for dynamics of $W_1$ and $W_2$ in gradient flow are

\begin{subequations}
\begin{align}
    \dot{W_1}&=-\frac{\partial\mathcal{L}}{\partial W_1}\label{eq:w1dot},\\\notag\\
    \dot{W_2}&=-\frac{\partial\mathcal{L}}{\partial W_2}\label{eq:w2dot}.
\end{align}
\end{subequations}

Using chain rule we have

\begin{subequations}
\begin{align}
    \frac{\partial\mathcal{L}}{\partial W_1}&=\frac{\partial\mathcal{L}}{\partial Z}\frac{\partial Z}{\partial W_1}\label{eq:partialw1},\\\notag\\
    \frac{\partial\mathcal{L}}{\partial W_2}&=\frac{\partial\mathcal{L}}{\partial \hat{X}}\frac{\partial \hat{X}}{\partial W_2}\label{eq:partialw2}.
\end{align}
\end{subequations}

We now evaluate each derivative:

\begin{subequations}
\begin{align}
    \frac{\partial\mathcal{L}}{\partial \hat{X}}&=\frac{1}{n}\left(\hat{X}-X\right)\notag\\\notag\\
    &=\frac{1}{n}\left(W_2W_1X-X\right)\\\notag\\
    \frac{\partial\mathcal{L}}{\partial Z}&=\frac{\partial\mathcal{L}_{rec}}{\partial Z}+\beta\frac{\partial\mathcal{L}_{norm}}{\partial Z}\notag\\\notag\\
    &=W_2^T\frac{\partial\mathcal{L}}{\partial \hat{X}}+\beta\left(\frac{1}{n}W_1X\right)\notag\\\notag\\
    &=\frac{1}{n}W_2^T\left(W_2W_1X-X\right)+\beta\left(\frac{1}{n}W_1X\right)\\\notag\\
    \frac{\partial Z}{\partial W_1}&=X^T\\\notag\\
    \frac{\partial \hat{X}}{\partial W_2}&=Z^T=X^TW_1^T.
\end{align}
\end{subequations}

Substituting in Eqs.~\ref{eq:partialw1} and~\ref{eq:partialw2} and considering Eqs.~\ref{eq:w1dot} and~\ref{eq:w2dot} we get

\begin{subequations}
\begin{align}
    \dot{W_1}&=-\frac{1}{n}W_2^T\left(W_2W_1X-X\right)X^T-\beta\left(\frac{1}{n}W_1XX^T\right),\\\notag\\
    \dot{W_2}&=-\frac{1}{n}\left(W_2W_1X-X\right)X^TW_1^T.
\end{align}
\end{subequations}

Substituting $\frac{1}{n}XX^T$ as $S$ and simplifying, we get

\begin{align}
    \dot{W}_1(t)&=[W_2^T-W_2^TW_2W_1-\beta W_1]S,\notag\\\notag\\
    \dot{W}_2(t)&=[I-W_2W_1]SW_1^T\notag.
\end{align}

\section{Architectures and Hyperparameters}\label{arch}

Below are the detailed architectures and dataset-specific hyperparameter values for training. 2D convolution and transposed convolution layers with channel size $c$, kernel size $k\times k$, and stride $s$ are denoted respectively as Conv($c$, $k$, $s$) and ConvT($c$, $k$, $s$). Leaky ReLU activation with negative slope $a$ is denoted as LReLU($a$). Linear layer with output size $l$ is denoted as Linear($l$)

\subsection{MNIST/Fashion-MNIST}

\noindent \textbf{Encoder:} Conv(64, 4, 1)-BatchNorm-LReLU(0.2)-Conv(128, 4, 1)-BatchNorm-LReLU(0.2)-Conv(256, 4, 2)-BatchNorm-LReLU(0.2)-Conv(512, 4, 1)-BatchNorm-LReLU(0.2)-Conv(64, 4, 1)-Flatten-Linear(128)/Linear(256).
\vspace{10pt}

\noindent \textbf{Decoder:} Linear(1024)-Unflatten(64, 4, 4)-ConvT(512, 4, 1)-BatchNorm-LReLU(0.2)-ConvT(256, 4, 1)-BatchNorm-LReLU(0.2)-ConvT(128, 4, 2)-BatchNorm-LReLU(0.2)-ConvT(64, 4, 1)-BatchNorm-LReLU(0.2)-ConvT(1, 4, 1).
\vspace{10pt}

\noindent $\beta=1000/100$\quad,\quad $lr=10^{-4}$.

\subsection{CIFAR-10/CIFAR-100}

\noindent \textbf{Encoder:} Conv(32, 5, 1, bias=False)-BatchNorm-LReLU(0.1)-Conv(64, 4, 2, bias=False)-BatchNorm-LReLU(0.1)-Conv(128, 4, 1, bias=False)-BatchNorm-LReLU(0.1)-Conv(256, 4, 2, bias=False)-BatchNorm-LReLU(0.1)-Conv(512, 4, 1, bias=False)-BatchNorm-LReLU(0.1)-Conv(512, 1, 1, bias=False)-BatchNorm-LReLU(0.1)-Conv(64, 1, 1)-Flatten.
\vspace{10pt}

\noindent \textbf{Decoder:} Unflatten(64, 1, 1)-ConvT(256, 4, 1, bias=False)-BatchNorm-LReLU(0.1)-ConvT(128, 4, 2, bias=False)-BatchNorm-LReLU(0.1)-ConvT(64, 4, 1, bias=False)-BatchNorm-LReLU(0.1)-ConvT(32, 4, 2, bias=False)-BatchNorm-LReLU(0.1)-ConvT(32, 5, 1, bias=False)-BatchNorm-LReLU(0.1)-Conv(32, 1, 1, bias=False)-BatchNorm-LReLU(0.1)-Conv(3, 1, 1).
\vspace{10pt}

\noindent $\beta=10/100$\quad,\quad $lr=10^{-2}/10^{-3}$.

\subsection{Arrhythmia}

\noindent \textbf{Encoder:} Linear(256)-BatchNorm-ReLU-Linear(128)-BatchNorm-ReLU-Linear(16).
\vspace{10pt}

\noindent \textbf{Decoder:} Linear(128)-BatchNorm-ReLU-Linear(256)-BatchNorm-ReLU-Linear(274).
\vspace{10pt}

\noindent $\beta=10$\quad,\quad $lr=10^{-3}$.

\end{document}